\documentclass{article}

\usepackage{microtype}
\usepackage{graphicx}
\usepackage{subfigure}
\usepackage{booktabs} 
\usepackage{makecell}
\usepackage{bbding}

\usepackage{tabularray}
\usepackage{amsmath}
\usepackage{amssymb}
\usepackage{booktabs}

\usepackage{epsfig}
\usepackage{pifont}
\usepackage{multirow}
\usepackage{algpseudocode}
\usepackage[table]{xcolor}
\usepackage{xcolor,colortbl}
\usepackage{bm}
\definecolor{mistygray}{rgb}{0.86, 0.86, 0.86}
\newcommand{\topresult}[1]{{\textbf{#1}}}
\newcommand{\secondresult}[1]{{\underline{#1}}}
\def\etal{\emph{et al}.}
\newcommand{\ie}{\emph{i.e.}}

\newcommand{\etc}{etc\@ifnextchar.{}{.\@}}

\usepackage[accepted]{icml2025}

\usepackage{amsmath}
\usepackage{amssymb}
\usepackage{mathtools}
\usepackage{amsthm}

\usepackage[capitalize,noabbrev]{cleveref}

\theoremstyle{plain}

\theoremstyle{definition}

\theoremstyle{remark}

\usepackage[textsize=tiny]{todonotes}

\icmltitlerunning{Efficient Motion Prompt Learning for Robust Visual Tracking}

\begin{document}

\twocolumn[
\icmltitle{Efficient Motion Prompt Learning for Robust Visual Tracking}

\begin{icmlauthorlist}
\icmlauthor{Jie Zhao}{a}
\icmlauthor{Xin Chen}{b}
\icmlauthor{Yongsheng Yuan}{a}
\icmlauthor{Michael Felsberg}{c}
\icmlauthor{Dong Wang}{a}
\icmlauthor{Huchuan Lu}{a}
\end{icmlauthorlist}

\icmlaffiliation{a}{Dalian University of Technology}
\icmlaffiliation{b}{City University of Hong Kong}
\icmlaffiliation{c}{Linköping University}

\icmlcorrespondingauthor{Dong Wang}{wdice@dlut.edu.cn}

\icmlkeywords{Machine Learning, ICML}

\vskip 0.3in
]



\printAffiliationsAndNotice{}  

\begin{abstract}
Due to the challenges of processing temporal information, most trackers depend solely on visual discriminability and overlook the unique temporal coherence of video data. 
In this paper, we propose a lightweight and plug-and-play motion prompt tracking method. It can be easily integrated into existing vision-based trackers to build a joint tracking framework leveraging both motion and vision cues, thereby achieving robust tracking through efficient prompt learning. 
A motion encoder with three different positional encodings is proposed to encode the long-term motion trajectory into the visual embedding space, while a fusion decoder and an adaptive weight mechanism are designed to dynamically fuse visual and motion features. 
We integrate our motion module into three different trackers with five models in total. Experiments on seven challenging tracking benchmarks demonstrate that the proposed motion module significantly improves the robustness of vision-based trackers, with minimal training costs and negligible speed sacrifice. Code is available at \url{https://github.com/zj5559/Motion-Prompt-Tracking}.
\end{abstract}

\section{Introduction}
\label{sec:intro}

\begin{figure}[tbp]
    \centering
    \includegraphics[width=\linewidth]{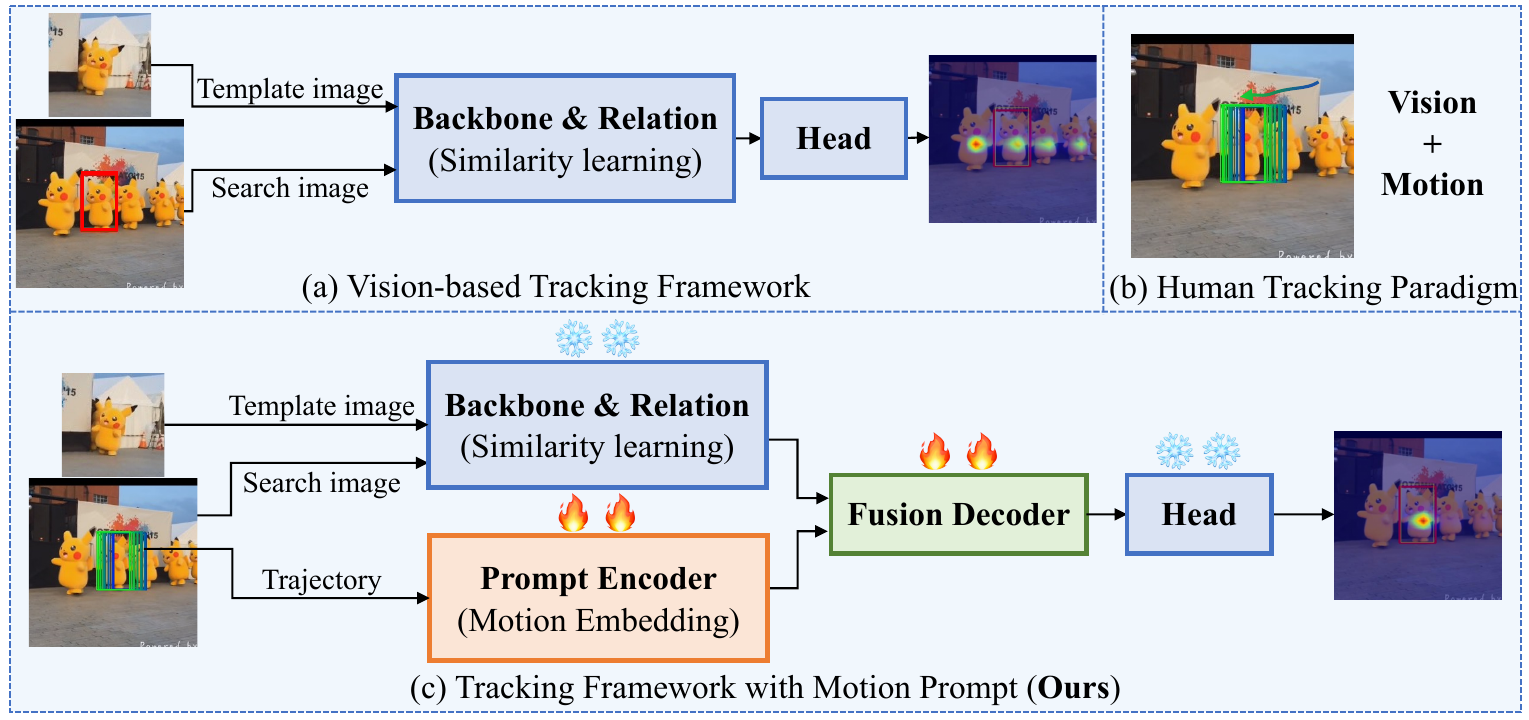}
        \caption{\textbf{Illustration of different tracking paradigms.}
    Our plug-and-play method enables visual trackers to benefit from motion prompts, making them more akin to the human tracking paradigm.}
    \vspace{-1.5em}
    \label{fig:prompt}
\end{figure}

Given a sequence and an arbitrary object with interest, visual object tracking (VOT) aims to locate this specified object in each subsequent frame. As one of the fundamental computer vision tasks, VOT has been developed rapidly over the past decade. 
Mainstream trackers~\cite{siamrpn++,dimp,zhao2022vision,chen2023seqtrack} perceive VOT as a visual matching problem between a pair of discrete image patches, including a template and a local search image cropped from the initial and current frames, respectively. These vision-based tracking frameworks mainly rely on the discriminative ability of visual models but overlook the crucial temporal coherence in videos. The temporal consistency of these trackers can be only reflected in determining the next local search region by the predicted target position. As shown in Fig.~\ref{fig:prompt}(a), due to inherent limitations of appearance cues, these vision-based trackers are prone to tracking drift when confronted with complex challenges, like distractors, severe occlusion, and so on.
Unlike relying solely on visual information, the human tracking paradigm~\cite{ramachandran1985neurobiology,sokhandan2024biological} seamlessly integrates both vision and motion cues. As shown in Fig.~\ref{fig:prompt}(b), humans can effortlessly identify the specified target among multiple visually similar objects by discerning its motion patterns from trajectories. 

However, due to the blend of object and camera movement, the irregularity in observed motion patterns makes it challenging to incorporate temporal coherence into tracking frameworks.
Recent state-of-the-art (SOTA) trackers~\cite{wei2023autoregressive, bai2023artrackv2,zheng2024odtrack} start exploring effective temporal mechanisms to enhance tracking accuracy, but sequential training~\cite{kim2022sequential} is required for these methods, which considerably increases computational and memory demands. Beyond the substantial training costs, constrained computing resources will also limit the ability of these models to perceive long temporal information. In this paper, we post a question: \textit{Is heavy sequential training necessary for trackers to capture temporal relations?} 
We answer this question by proposing a training-efficient \textbf{M}otion \textbf{P}rompt-based \textbf{T}racking module (MPT), which demonstrates that sequential training is unnecessary for temporal-related tracking.

Specifically, our MPT module can be flexibly integrated into existing vision-based trackers to achieve joint tracking based on both visual and motion cues. Within a lightweight frame-level fine-tuning, our method can enhance various baseline trackers to achieve comparable performance to those sequential-based SOTA methods.
As shown in Fig.~\ref{fig:prompt}(c), our joint tracking framework contains a fixed vision-based tracker and several additionally proposed motion modules. The prompt encoder (the orange block) independently encodes historical trajectories into the visual embedding space by three different positional encodings.
A fusion decoder (the green block) is then employed to dynamically fuse visual and motion features by a Transformer network and an adaptive weighting mechanism.
Furthermore, we adopt prompt learning to integrate our motion modules into vision-based SOTA trackers more efficiently and flexibly. We develop our MPT as a plug-and-play prompt module, aiming to learn the ability for dynamic adaptation to baseline visual trackers. All parameters of the baseline are fixed during training, enabling our method to have fast training with minimal memory requirements.
By integrating motion prompts from object trajectories, our MPT can sustain robust tracking under challenging scenarios, reflecting a tracking paradigm more akin to human-like behavior.

Our main contributions can be summarized as follows:
\begin{itemize}
    \item We propose a plug-and-play motion prompt tracking module (MPT), which can be flexibly integrated into various tracking models to capture and utilize temporal coherence from historical trajectories. 
    \item Efficient prompt learning is employed to fine-tune our method, which unbinds the necessity of heavy sequential training for temporal learning, thereby releasing huge requirements of training resources.
    \item We integrate our MPT into three visual trackers across five models. Experiments on seven benchmarks demonstrate that our MPT enhances the robustness of existing vision-based trackers in challenging scenarios, within minimal training costs and negligible speed sacrifice.
\end{itemize}

\section{Related Work}
\label{sec:related_work}

In this section, we review mainstream visual trackers, temporal-related trackers, and prompt learning, respectively.

\noindent\textbf{Visual Tracking.}
Existing trackers commonly adopt a visual matching strategy to model the tracking task. 
Most earlier trackers~\cite{SiamFC, ASRCF, siamrpn++} employ correlation operations to compute the similarity map between visual features of the template and search region. 
TransT~\cite{chen2021transt} and subsequent Transformer-based works~\cite{ye2022ostrack,swintrack,chen2023seqtrack} employ Transformer networks~\cite{vaswani2017attention,dosovitskiy2021vit} to obtain the fused features of the template and search region, which is then used to make the final prediction. Due to global correlation and deep fusion, these trackers lead to improved accuracy compared to similarity-based methods.
Despite achieving superior performance on existing benchmarks, such solely vision-based matching strategies encounter difficulties in complex visual environments. To alleviate this bottleneck, our work proposes a general tracking module based on motion prompts. By incorporating temporal motion cues into these vision-based trackers, we enrich the information available to the head network for predictions, thereby significantly improving the tracking robustness in challenging scenarios.

\begin{figure*}[!htbp]
    \centering
    \includegraphics[width=0.9\linewidth]{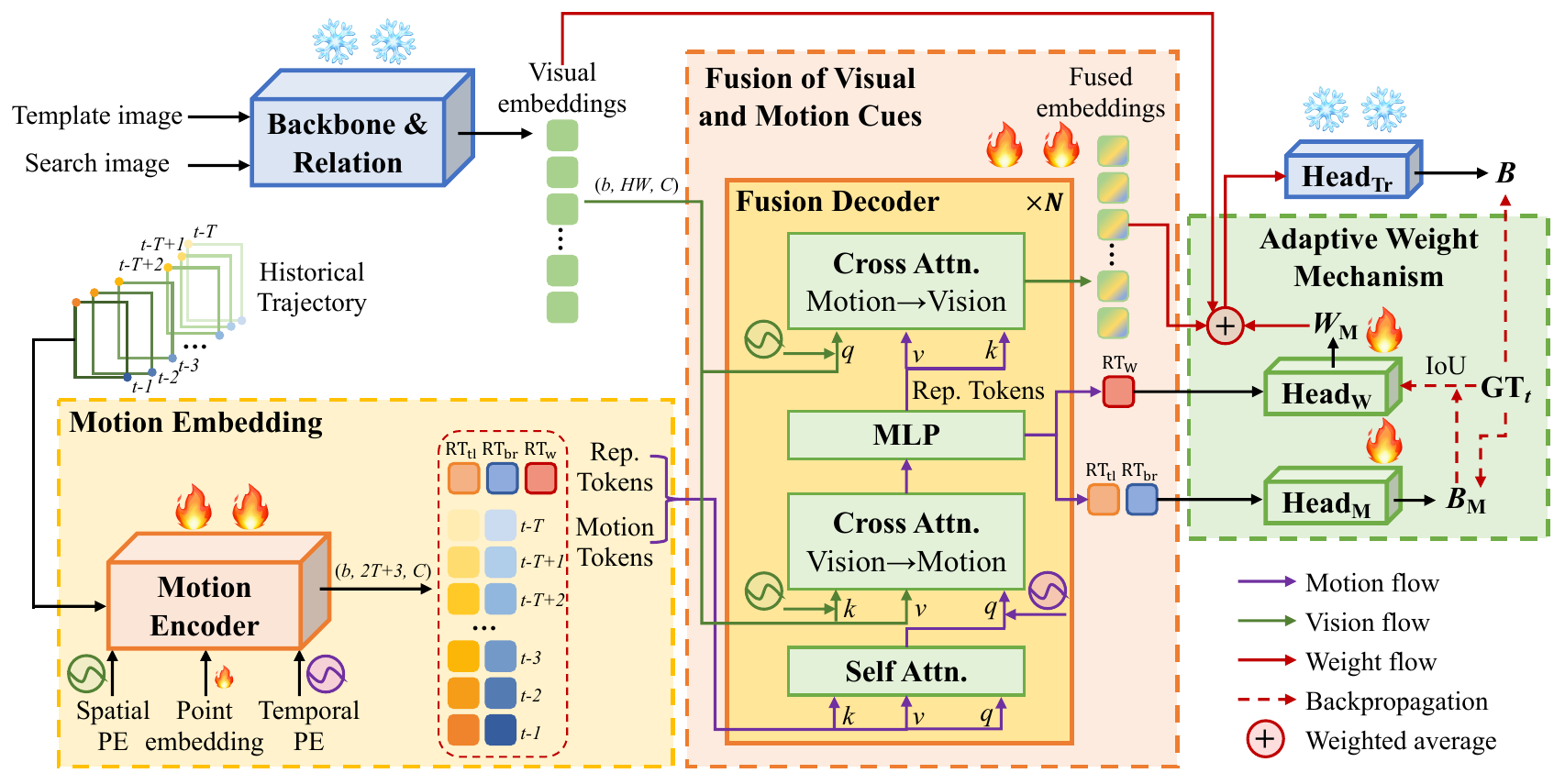}
    \caption{\textbf{Pipeline of our motion prompt-based tracking method.} The historical trajectory is first embedded into the visual embedding space by our motion encoder, and then fused with the visual embedding using the proposed fusion decoder. An adaptive weight mechanism is employed to further dynamically adjust vision and motion cues. Ultimately, the obtained fused embedding is used to make a robust tracking prediction by the tracking head $\rm Head_{Tr}$.} 
    \vspace{-0.5em}
    \label{fig:framework}
\end{figure*}

\noindent\textbf{Temporal-related Tracking.}
Beyond visual information, some works have investigated the significance of temporal information for robust tracking. Some traditional trackers~\cite{weng2006video,zhang2014new} primarily utilize kalman filtering to predict the next motion step of the target. However, they assume that the target undergoes regular motion patterns, which usually do not align with the complex tracking scenarios. The lack of deep visual representations also leads to a substantial lag in their performance.

In the era of deep learning, some methods~\cite{dimp,zhao2022robust,yan2021stark,cui2022mixformer,liu2024spatial} employ update mechanisms for models or templates to capture the temporal changes of visual information, these methods primarily depend on richer appearance information to enhance tracking. TCTrack++~\cite{cao2023TCTrack} and ODTrack~\cite{zheng2024odtrack} employ heavy temporal mechanisms, which implicitly integrate temporal information via propagating consecutive visual features. 
Different from them, our lightweight module directly encodes and integrates motion trajectories into the visual tracking framework, offering a fresh perspective. 
Our experiments with the template update tracker SeqTrack~\cite{chen2023seqtrack} also demonstrate that benefits derived from motion and updates are complementary rather than conflicting.

Recently, Alireza~\etal~\cite{sokhandan2024biological} adopt optical flows to represent motion vectors, which raises computational burdens. SLT~\cite{kim2022sequential} proposes a sequential training method, incorporating motion cues into training by locating short video clips. 
But it still follows vision-based tracking in inference. Implicitly learning temporal correlations among consecutive frames is also a complex and computationally intensive task. 
ARTrack~\cite{wei2023autoregressive} and its extension~\cite{bai2023artrackv2} share a similar motivation with ours, integrating temporal information into tracking by individually encoding the motion trajectory. However, its motion encoding is tailored for autoregressive models, which limits its applicability to other tracking frameworks. Autoregressive model and sequential training also increase memory and computational costs. Limited hardware memory will further restrict the model from perceiving long trajectories. In comparison, our method employs a low-cost prompt learning approach to leverage longer motion trajectories for guiding existing trackers, offering greater flexibility and versatility.

\noindent\textbf{Prompt Learning.}
Prompt learning is an emerging machine learning paradigm, garnering particular attention in the field of natural language processing~\cite{liu2021pre}. Its core idea involves interacting with the model via task-related prompts rather than directly adjusting model parameters. In computer vision and multi-modal domains, prompt learning proves to be more efficient and flexible than traditional fine-tuning methods. It shows success in various tasks, including visual recognition~\cite{vpt}, vision-language understanding~\cite{Visual_instruction_tuning}, dense prediction~\cite{adapter_dense}, and video understanding~\cite{video_prompt}. 

In the field of visual tracking, in addition to the aforementioned ARTrack and its extension, which adopt trajectories as prompts by employing sequential training and autoregressive models, ProTrack~\cite{protrack} and ViPT~\cite{VIPT} utilize prompt learning to achieve superior performance for multi-modal tracking, yet they do not exploit the potential of prompt learning in general visual tracking.
PromptVT~\cite{zhang2024promptvt} can be treated as an appearance-based prompt method. It efficiently enhances appearance features using dynamic appearance information, thereby enabling the model to be robust to appearance changes during tracking. However, the tracking problem is still treated as a discrete image processing task. In contrast, our MPT leverages the temporal consistency of videos by incorporating continuous object trajectories into the visual tracking framework, offering a new perspective on employing prompt learning in the tracking field.
Our method proves to be more effective in addressing visually challenging scenarios, such as occlusions and distractors. 



%

\section{Method}

\subsection{Revisit of Vision-based Tracking Framework}
Mainstream trackers formulate the tracking problem as 
\begin{align}
    \centering
    \label{eq:pipeline1}
    B^t &={\rm Head_{Tr}}(\varphi_{v}(Z,X^t)).
\end{align}
A visual encoder $\varphi_{v}(\cdot)$ is commonly employed to embed both the template image $Z$ and the search image $X^t$, producing the discriminative visual representations, like similarity maps~\cite{siamrpn++,dimp} or fused features~\cite{cui2022mixformer,chen2022simtrack} of image pairs. The final tracking result $B^t$ for the $t$-th frame is predicted by a tracking head module, $\rm Head_{Tr}$. Some trackers~\cite{ye2022ostrack,yan2021stark} employ a lightweight head to directly predict coordinates, whereas other methods~\cite{chen2023seqtrack, wei2023autoregressive} implement the head module as a heavy decoder to tackle the sequential coordinate prediction problem. For these vision-based tracking methods, a powerful visual encoder $\varphi_{v}$ is necessary to achieve robust tracking. However, good visual models need to achieve invariant representations to appearance changes for the same object, while maintaining distinctive for distractors. Acquiring such balanced discriminative visual representations poses a challenging learning task for visual models. In addition, when faced with occlusion, illumination changes, or distractors, visual information often turns out to be unreliable, leading to tracking drift in these vision-based trackers.

In this paper, we propose a motion prompt-based tracking method, which can integrate motion cues into vision-based baseline trackers by prompt learning. This joint tracking framework incorporating both motion and vision cues can adaptively extract complementary features, thereby enhancing the robustness of tracking predictions, particularly in the face of complex challenges.

\subsection{Overview of Motion Prompt Tracking}
Our tracking method with motion prompts is formulated as
\begin{align}
    \centering
    \label{eq:pipeline2}
    B^t &={\rm Head_{Tr}}(\boldsymbol{{\mathcal{D}_{f}}}(\varphi_{v}(Z,X^t),\boldsymbol{\varphi_{m}}(B^{t-T:t-1}))).
\end{align}
In addition to the visual encoder $\varphi_{v}(\cdot)$, we employ a motion encoder $\varphi_{m}(\cdot)$ to embed the historical trajectory $B^{t-T:t-1}$ into the latent embedding space consistent with visual features. The trajectory is denoted as consecutive bounding boxes of the target from previous $T$ frames. A decoder $\mathcal{D}_{f}(\cdot)$ is then utilized to fuse motion and visual features, and the resulting fused representation will be forwarded to the downstream head to generate robust tracking predictions. 


As shown in Fig.~\ref{fig:framework}, the two input flows, including the visual input and the object motion trajectory, are processed separately through the baseline visual model and our motion encoder to extract corresponding features. 
Through calculations of several self-attention and cross-attention modules, features from the two branches are complementarily fused. An adaptive weight mechanism is additionally utilized to establish a dynamic residual connection between visual features and the output of the fusion decoder, thereby achieving stable fused embeddings. 
By integrating complementary motion prompts, the proposed motion modules empower the vision-based baseline tracker to achieve enhanced performance, especially under challenging scenes.

\subsection{Model Architecture}
Our motion prompt tracking method mainly consists of three modules, \ie, a motion encoder, a fusion decoder, and an adaptive weight mechanism. Detailed architectures of each module are described as follows.

\noindent\textbf{Motion Encoder.}
Given consecutive object bounding boxes of previous $T$ frames, our motion encoder embeds these sequential coordinates into multiple motion tokens aligned with the visual embedding space. The following three types of positional encodings (PE) are utilized to characterize each token across both spatial and temporal dimensions.

\textit{(1) Spatial positional encoding.} Inspired by SAM~\cite{kirillov2023segment}, we represent each bounding box using the normalized coordinates of two corner points, namely, the top-left and bottom-right points. A Gaussian-based spatial PE~\cite{tancik2020fourier} is adopted to map 2-dimensional point coordinates into $C$-dimensional spatial vectors, aligning the motion embedding with the visual embedding in terms of dimensions.

\textit{(2) Point embedding.} After being mapped by the spatial PE, the trajectory of $T$ frames is transformed to a group of motion tokens $\in \mathbb{R}^{2T\times C}$. To distinguish between the two corner point types for each token, we introduce two learnable point embeddings and add them to their respective motion tokens. This enables the model to perceive the point type of each motion token.

\textit{(3) Temporal positional encoding.} To establish temporal orders for motion tokens, we incorporate a learnable temporal PE into each token. Tokens from the same frame share identical temporal PE. Since the motion information closer to the current frame is typically more crucial in a long-term trajectory, we initialize this temporal PE in a non-linear manner, formulated as follows:
\begin{equation}
\label{eq:temproal_pe}
    \begin{split}
    {\rm PE}(t,2i) &=\sin\left(\frac{\alpha \lg(t+1)}{n^{2i/d}}\right), \\
    {\rm PE}(t,2i+1) &=\cos\left(\frac{\alpha \lg(t+1)}{n^{2i/d}}\right).
    \end{split}
\end{equation}
Here, $i, d$ denotes the dimension variable and the total dimension, $t \in [0, T-1]$ represents the temporal position of the point in the trajectory. A larger $t$ corresponds to a closer position to the current frame. The constant $n$ is set to $10^4$, and $\alpha$ is a parameter employed to control the frequency range.
To maximize spatial resolution without causing aliasing, we set $\alpha=7.23$ according to the Nyquist frequency~\cite{nyquist}. Both theoretical proof and experimental analysis can be found in Appendix~\ref{sec:nontpe}. 

In addition to the $2T$ motion tokens, we also introduce three learnable representative (Rep.) tokens. Among them, $\rm RT_{tl}$ and $\rm RT_{br}$ are employed to summarize motion information from all top-left and bottom-right trajectory points, respectively. $\rm RT_{w}$ represents the confidence of the motion prompt. We set two sub-tasks to guide the learning of three Rep. tokens, namely, bounding box regression and weight regression. These sub-tasks encourage our model to develop the ability to extract crucial motion prompts and predict dynamic weights.
As the output of the motion encoder, we concatenate $2T$ motion tokens and the three Rep. tokens to represent the extracted motion feature.

\noindent\textbf{Fusion Decoder.} 
As shown in Fig.~\ref{fig:framework}, we propose a lightweight fusion decoder to fuse the visual feature $f_v \in \mathbb{R}^{b\times HW\times C}$ and motion feature $f_m \in \mathbb{R}^{b\times (2T+3)\times C}$, where $b$ denotes the batch size, and $H, W$ represent resolutions of the visual feature.
The proposed fusion decoder consists of $N$ Transformer-based blocks, each including a motion self-attention module, two cross-attention modules with different directions, and a multi-layer perception (MLP) layer. 
Specifically, the self-attention module first explores crucial motion information from motion features themselves. 
The first cross-attention module then fuses visual cues into motion features, in which motion features $f_m$ are set as queries ($q$). After the update by an MLP, we take the three Rep. tokens as keys ($k$) and values ($v$) for the second cross-attention module, and update the visual feature $f_v$ (as queries) with representative motion prompts. 
For each cross-attention module, we incorporate a dense spatial map as positional encoding onto visual features. This ensures a correspondence between visual and motion features in spatial coordinates. Regarding temporal positions, we also apply the temporal PE to motion features in the first cross-attention module.

\noindent\textbf{Adaptive Weight Mechanism.}
Considering the instability of motion trajectories, we calculate the weighted average of the fused feature and original visual feature using an adaptive weight mechanism.
A weight head $\rm Head_W$ is employed to dynamically predict the weight $W_{\rm M}$ according to the output $\rm RT_{w}$ from MLP. 
To explicitly guide the learning of $W_{\rm M}$, we additionally introduce a motion head $\rm Head_M$ to predict the current bounding box $B_{\rm M}$ according to $\rm RT_{tl}$ and $\rm RT_{br}$. We calculate the Intersection-over-Union (IoU) of the groundtruth and $B_{\rm M}$, and use it to explicitly supervise the learning for $W_{\rm M}$. 





\subsection{Training and Inference}
\label{sec:train}
\noindent \textbf{Training.}
We adopt prompt learning to achieve efficient training. All parameters in the baseline tracker are fixed, and only our MPT modules are fine-tuned, including the motion encoder, fusion decoder, learnable embeddings, $\rm Head_W$, and $\rm Head_M$.
The backpropagation process is illustrated as dashed red arrows in Fig.~\ref{fig:framework}, The loss function for our MPT comprises the following three terms:
\begin{itemize}
    \item \textit{Baseline tracking loss.} We follow the same tracking loss as baseline trackers, namely $\mathcal{L}_{\rm Tr}(B, {{\rm GT}_t})$, where $B$ denotes the output of $\rm Head_{Tr}$, and ${\rm GT}_t$ represents the groundtruth of the object bounding box.
    
    \item \textit{Motion regression loss.} To enable Rep. tokens $\rm RT_{tl}$ and $\rm RT_{br}$ to extract essential motion patterns, we set a bounding box regression task on $\rm Head_M$. The generalized IoU loss~\cite{rezatofighi2019generalized} and ${\ell_1}$ loss are employed to supervise this sub-task, formulated as
        \begin{equation}
            \mathcal{L}_{\rm M}=\lambda_{\rm IoU}\mathcal{L}_{\rm IoU}+\lambda_{\ell_1}\mathcal{L}_{1},
        \end{equation}
    where $\lambda_{\rm IoU}=2$ and $\lambda_{\ell_1}=5$ in our experiments.

    \item \textit{Adaptive weight loss.} In our assumption, a lower quality of the motion tracking prediction $B_{\rm M}$ typically implies less confidence in motion prompts. We thereby adopt the IoU between $B_{\rm M}$ and ${\rm GT}_t$ as the label of the predicted weight $W_{\rm M}$, and calculate MSE loss to achieve explicit supervision. The adaptive weight loss is formulated as
        \begin{equation}
            \mathcal{L}_{\rm W}=\mathcal{L}_{\rm MSE}(W_M, {\rm IoU}(B_{\rm M}, {\rm GT}_t)).
        \end{equation}
\end{itemize}

\noindent The overall loss function of our MPT involves the above three terms, represented as 
\begin{equation}
    \mathcal{L}=\mathcal{L}_{\rm Tr}+\lambda_{\rm M}(\mathcal{L}_{\rm M}+\mathcal{L}_{\rm W}),
\end{equation}
where $\lambda_{\rm M}=1$ in our experiments.

\noindent \textbf{Inference.} During the inference, predicted bounding boxes of previous $T$ frames are first aligned and normalized based on the coordinate system of the current search region. Normalized trajectory is sent together with the image pair into our motion prompt-based tracking model for tracking prediction. As $\rm Head_M$ primarily serves to facilitate the more effective training for our motion modules, this network is disregarded during inference to save computational costs.

\section{Experiment}
To demonstrate the effectiveness and training efficiency of our MPT, we integrate our MPT into two different visual trackers, \ie, a one-stream tracker OSTrack\footnote{The official model without the candidate elimination module.}, and an autoregressive-based tracker SeqTrack. We also apply our MPT to ARTrack to show that our MPT can further complement temporal-related trackers rather than conflicting. We select a total of five baseline models with different backbones and resolutions. To distinguish them, base / large backbones are represented as ``B / L", while the following numbers represent image resolutions. We compare our methods with their baselines and other SOTA trackers on seven challenging tracking benchmarks. Experimental results demonstrate that our MPT can be seamlessly integrated into various trackers, enhancing their tracking robustness with fast and memory-efficient training.

\subsection{Implementation Details}
Our methods are implemented in Python with PyTorch. Models are trained on 2 NVIDIA A100 GPUs, and tested on a single NVIDIA RTX2080Ti GPU.

\noindent \textbf{Model.} The length of the historical trajectory $T$ is set to 30 based on experimental results. Each 2-dimensional coordinate is encoded into a motion token with the same dimension as baseline visual features. The lightweight fusion decoder is implemented as a two-layer Transformer network. 
The weight head $\rm Head_W$ and motion head $\rm Head_M$ are implemented by a two-layer MLP, where the hidden size is 256. 

\noindent \textbf{Training.} We adopt prompt learning to efficiently train our MPT, where the baseline model is frozen. 
We select the training splits of LaSOT~\cite{lasot}, GOT-10K~\cite{huang2019got10k}\footnote{Following the VOT protocol, 1k sequences are removed.}, and TrackingNet~\cite{trackingnet} as the training data.
For the motion input, we adopt DiMP-18~\cite{dimp} to generate real tracking predictions for each of training sequences, and employ reverse sampling, sparse sampling and CutMix~\cite{yun2019cutmix} for data augmentations.
More implementation details can be found in Appendix~\ref{sec:detail}, and extended experimental analysis in terms of data augmentations and trajectory length can be found in Appendix~\ref{sec:ext_ablation}.

\begin{table}[tbp]
\centering
\small
\caption{\textbf{Performance comparison on three VOT datasets.} The best results are highlighted with \topresult{bold}.}
\label{tab:vot}
\resizebox{\linewidth}{!}{
\setlength{\tabcolsep}{0.5mm}{
\begin{tabular}{c|cc|cc|cc}
\toprule[1.5pt]
 \multirow{2}{*}{Method} &\multicolumn{2}{c|}{VOT2018} &\multicolumn{2}{c|}{VOT2020}  &   \multicolumn{2}{c}{VOT2022 (STB)} \\ 
    & EAO $(\uparrow)$  &  R $(\downarrow)$  & EAO $(\uparrow)$ &  R $(\uparrow)$ & EAO $(\uparrow)$ &  R $(\uparrow)$  \\ \midrule[1.5pt]
MixFormer-22k&0.228&1.836 &0.317&0.824 &0.533&0.844\\
ROMTrack-B256&0.357&1.010 &0.318&0.815 &0.554&0.825\\
\hline
OSTrack-B256&0.287&1.204 &0.311&0.803&0.530&0.823\\
$+$\textbf{MPT (Ours)} &0.406&0.473 &0.334&\topresult{0.846}&0.572&0.859\\
\hline
SeqTrack-B256&0.339&1.067  &0.316&0.806&0.523&0.815\\
$+$\textbf{MPT (Ours)} &\topresult{0.428}&\topresult{0.471} &0.315&0.810&0.528&0.825\\ 
\hline
ARTrack-B256&0.399  & 0.599 &0.315  & 0.809 &0.532  & 0.818\\ 
$+$\textbf{MPT (Ours)}&0.469&0.491&\topresult{0.336}&\topresult{0.828}&\topresult{0.566}&\topresult{0.842}\\ 
\midrule[1.5pt]
MixFormer-L &0.238&1.596 &0.325&0.825 &0.549&0.843\\
ROMTrack-B384 &0.297&1.231 &0.309&0.794 &0.540&0.812\\
ARTrack-L384&0.391&0.520&0.336&0.834&0.570&0.852\\
\hline
OSTrack-B384&0.271&1.408 &0.288&0.767&0.518&0.796\\
$+$\textbf{MPT (Ours)} &0.363&0.679  &0.314&0.819&0.548&0.826\\
\hline
SeqTrack-L384&0.319&1.117 &0.337&0.859 &0.568&0.869\\
$+$\textbf{MPT (Ours)}&\topresult{0.461}&\topresult{0.414}  &\topresult{0.341}&\topresult{0.870} &\topresult{0.579}&\topresult{0.882}\\ 

\midrule[1.5pt]
Average gain&$+10.2\%$&$+57.3\%$ &$+1.5\%$&$+2.6\%$ &$+2.4\%$&$+2.3\%$\\
\bottomrule[1.5pt]
\end{tabular}
}}
\vspace{-1em}
\end{table}
\begin{table*}[!tb]
\centering
\caption{\textbf{Comparisons on four tracking benchmarks.} Top two results (except for \colorbox{mistygray!50}{sequential-training methods}) are marked in \topresult{bold} and \secondresult{underlined}. For a fair comparison, we list the training strategy (Train) and template amounts (Temp.) of each tracker.}

\label{tab:sota}
\resizebox{\linewidth}{!}{%
\setlength{\tabcolsep}{0.5mm}{
\begin{tabular}{c|cc|llc|llc|llc|llc}
\toprule[1.5pt]
\multirow{2}{*}{Method}  &\multirow{2}{*}{Train}&\multirow{2}{*}{Temp.}  & \multicolumn{3}{c|}{LaSOT} & \multicolumn{3}{c|}{$\rm LaSOT_{\rm EXT}$} &  \multicolumn{3}{c|}{TNL2K} & \multicolumn{3}{c}{TrackingNet} \\
                & &&    AUC  & $\rm P_{norm}$ & Pre  & AUC     & $\rm P_{norm}$    & Pre     & AUC  & $\rm P_{norm}$ & Pre & AUC  & $\rm P_{norm}$ & Pre   \\ \midrule[1.5pt]
                 
SiamRCNN~\cite{voigtlaender2020siamrcnn}&Frame&1&64.8  & 72.2 &-&-& - & - &- &-&-&81.2&85.4&80.0 \\
TransT~\cite{chen2021transt}&Frame&1& 64.9 & 73.8     & 69.0 & -       & -           & -       &50.7 &-&51.7&81.4&86.7&80.3   \\
STARK-ST101~\cite{yan2021stark}&Frame&2&67.1&77.0&-&-&-&-&-&-&-&82.0&86.9&-\\ 
ToMP-101~\cite{mayer2022transforming}&Frame&1& 68.5& 79.2 & 73.5 & 45.9 & - &- &-&-&-&81.5&86.4&78.9  \\
AiATrack~\cite{gao2022aiatrack}&Frame&2& 69.0 & 79.4 &73.8 &46.8  & 54.4& 54.2 &-&-&-&82.7&87.8&80.4  \\
MixFormer-22k~\cite{cui2022mixformer}&Frame&2&69.2&68.7&74.7&   -  & -& - &-&-&-&83.1&88.1&81.6\\
Sim-B16~\cite{chen2022simtrack}&Frame&1&69.3&78.5&-&-&-&-&54.8&-&53.8&82.3&86.5&-\\
ROMTrack-B256~\cite{chen2022simtrack}&Frame&2& 69.3 & 78.8 &75.6 &48.9  & 59.3& 55.0 &-&-&-&83.6&88.4&82.7  \\

MixViT (ConvMAE)~\cite{mixformer2024}&Frame&2&\topresult{70.4}&\topresult{80.4}&\secondresult{76.7}&-&-&-&-&-&-&\topresult{84.5}&\topresult{89.1}&\topresult{83.7}\\
\hline
OSTrack-B256~\cite{ye2022ostrack}&Frame&1&68.6&78.0&74.5 &46.9&56.9&52.8 &55.6&72.1&56.1 &83.0&87.7&81.7 \\
+\textbf{MPT (Ours)} &Frame&1&68.9&78.6&74.8 &48.7&59.1&55.2 &55.8&72.7&56.5 &83.6&88.4&82.4 \\
\hline
SeqTrack-B256~\cite{chen2023seqtrack}&Frame&2&69.4&79.2&75.8 &49.8&\secondresult{61.2}&56.8 &56.5&74.2&58.6 &83.2&88.3&82.3 \\
+\textbf{MPT (Ours)} &Frame&2&70.1&\topresult{80.4}&\secondresult{76.7} &\topresult{50.8}&\topresult{62.4}&\topresult{58.0} &57.8&\topresult{75.9}&\topresult{60.3} &83.6&88.8&82.6 \\
\hline
ARTrack-B256~\cite{wei2023autoregressive}&\textbf{Seq}&1&\topresult{70.4} & 79.5 &76.6&48.4&57.7&53.7&\secondresult{57.9}&74.0&59.6&\secondresult{84.2}&88.7&\secondresult{83.5}\\
+\textbf{MPT (Ours)}&Frame&1&70.3&80.1&\topresult{76.8}&\secondresult{50.4}&60.7&\secondresult{57.5}&\topresult{58.5}&\secondresult{75.2}&\topresult{60.9}&\secondresult{84.2}&\secondresult{88.9}&83.2 \\ 
\hline
\rowcolor{mistygray!50}ARTrackV2-B256~\cite{bai2023artrackv2}&\textbf{Seq}&2&71.6&80.2&77.2&50.8&61.9&57.7&59.2&-&-&84.9&89.3&84.5\\
\midrule[1.5pt]

MixFormer-L~\cite{cui2022mixformer}&Frame&2& 70.1 & 79.9 &76.3 &-  & -& - &-&-&-&83.9&88.9&83.1  \\
Sim-L\/14~\cite{chen2022simtrack}&Frame&1& 70.5 & 79.7 &- &-  & -& - &55.6&-&55.7&83.4&87.4&-  \\
ROMTrack-B384~\cite{cai2023romtrack}&Frame&2& 71.4 & 81.4 &78.2 &51.3  & 62.4& 58.6 &-&-&-&84.1&89.0&83.7  \\
MixViT-L (ConvMAE)~\cite{mixformer2024}&Frame&2&\secondresult{73.3}&\secondresult{82.8}&\secondresult{80.3}&-&-&-&-&-&-&\topresult{86.1}&\topresult{90.3}&\secondresult{86.0}\\
\hline
OSTrack-B384~\cite{ye2022ostrack}&Frame&1&71.0&80.9&77.3 &50.9&61.6&57.7 &57.5&74.2&58.6 &83.8&88.4&83.1 \\
+\textbf{MPT (Ours)} &Frame&1&70.7&80.5&76.9 &\topresult{52.8}&\topresult{64.0}&\topresult{60.7} &57.8&74.8&59.5 &84.6&89.1&83.8 \\
\hline
SeqTrack-L384~\cite{chen2023seqtrack}&Frame&2&72.5&81.4&79.2 &50.4&61.2&57.2 &\secondresult{59.6}&\secondresult{76.6}&\secondresult{63.5} &85.6&90.0&85.9 \\
+\textbf{MPT (Ours)} &Frame&2&\topresult{73.9}&\topresult{83.2}&\topresult{81.0} &\secondresult{51.7}&\secondresult{62.7}&\secondresult{58.8} &\topresult{60.4}&\topresult{77.5}&\topresult{64.2} &\secondresult{85.8}&\secondresult{90.1}&\topresult{86.2} \\
\hline

\rowcolor{mistygray!50}ARTrack-L384~\cite{wei2023autoregressive}&\textbf{Seq}&1&73.1&82.0&80.2&52.0&62.4&59.0&60.8&77.5&64.5&85.4&90.2&85.8 \\
\rowcolor{mistygray!50}ODTrack-B384~\cite{zheng2024odtrack}&\textbf{Seq}&3& 73.2 & 83.2 &80.6 &52.4  & 63.9& 60.1 &60.9&-&-&85.1&90.1&84.9  \\
\rowcolor{mistygray!50}ARTrackV2-L384~\cite{bai2023artrackv2}&\textbf{Seq}&2&73.6&82.8&81.1&53.4&63.7&60.2&61.6&-&-&86.1&90.4&86.2 \\
\rowcolor{mistygray!50}ODTrack-L384~\cite{zheng2024odtrack}&\textbf{Seq}&3& 74.0 & 84.2 &82.3 &53.9  & 65.4& 61.7 &61.7&-&-&86.1&91.0&86.7  \\

\midrule[1.5pt]
Average gain&-&-&$+0.4\%$&$+0.8\%$&$+0.6\%$ &$+1.6\%$&$+2.1\%$&$+2.4\%$ &$+0.6\%$&$+1.0\%$&$+1.0\%$ &$+0.4\%$&$+0.4\%$&$+0.3\%$\\
 \bottomrule[1.5pt]
\end{tabular}
}}
\vspace{-1.0em}
\end{table*}

\subsection{State-of-the-Art Comparison}
We compare our methods with baselines and other SOTA trackers on the following seven tracking benchmarks. 

\noindent \textbf{VOT.} The committee of VOT challenge proposes a series of challenging tracking benchmarks. Trackers are evaluated primarily by the expected average overlap (EAO), which is a principled combination in terms of tracking accuracy and robustness (R). As shown in Tab.~\ref{tab:vot}, we compare our methods with several SOTA trackers on three representative VOT datasets, including VOT2018~\cite{vot2018}, VOT2020~\cite{vot2020}, and VOT2022~\cite{vot2022}. For a fair comparison, all of the presented trackers are run with their official models and settings, and evaluated based on bounding box outputs. 
Compared with SOTA methods, trackers integrating our MPT exhibit superior performance on these challenging sequences, especially in terms of robustness. Our MPT significantly enhances the performance of the five baseline models, averaging an improvement of 2.0\% EAO across VOT2020 and VOT2022, and 10.2\% EAO on VOT2018. 

\noindent \textbf{LaSOT and} $\rm \mathbf{LaSOT_{EXT}}$. LaSOT~\cite{lasot} contains 280 long-term sequences with 70 different categories, while its extension $\rm LaSOT_{EXT}$~\cite{fan2021lasotext}, consists of 150 extremely challenging sequences with 15 unseen categories. The success rate (AUC), normalized precision ($\rm P_{norm}$), and precision (Pre) are adopted to evaluate trackers. As reported in Tab.~\ref{tab:sota}, our MPT increases baseline trackers by an average 1.6\% AUC on $\rm LaSOT_{EXT}$. Furthermore, our training-efficient method (\textbf{+MPT}) enhances SeqTrack-L384 to achieve an AUC of 73.9\% on LaSOT, which is competitive with sequential-level training method using multi-template strategy, ODTrack~\cite{zheng2024odtrack}.

\noindent \textbf{TNL2K~\cite{wang2021tnl2k}.} TNL2K is a recently released dataset with 700 challenging sequences. As shown in Tab.~\ref{tab:sota}, our MPT yields average 0.6\% and 1.0\% improvements in terms of AUC and $\rm P_{norm}$, respectively.

\noindent \textbf{TrackingNet~\cite{trackingnet}.} TrackingNet contains 511 sequences covering diverse categories and scenarios. As presented in Tab.~\ref{tab:sota}, our MPT brings baselines an average performance improvement of around 0.4\% on each metric.

\noindent \textbf{Discussion.} We find that the performance improvement of our MPT is slightly biased across different datasets, with more substantial gains on more challenging ones (like VOTs, $\rm LaSOT_{EXT}$). To explore the underlying reasons, we divide each of the three datasets (\ie, LaSOT, $\rm LaSOT_{EXT}$, and TNL2K) into ``hard / easy" subsets based on the performance of each sequence, and analyze the performance changes of MPT on each subset (shown as Tab.~\ref{tab:bias}). As discussed in Appendix~\ref{sec:ext_ablation}, our MPT shows great consistency in hard / easy sets, achieving average AUC changes of +4.4\% / -0.8\%, with small standard deviations of 0.6 / 0.4 across five baselines. Therefore, we reach a consistent conclusion that the advantage of our MPT is more evident in hard scenarios. This is because our MPT primarily aims to reduce tracking failures by motion cues, leading to a significant improvement in tracking robustness rather than accuracy. This can also be intuitively demonstrated by the performance gains on the three VOT benchmarks (shown as Tab.~\ref{tab:vot}).

\begin{table}[htbp]
    \caption{\textbf{Efficiency of training and inference}. Two A100 GPUs are used with 128 batch sizes (BS), except for ARTrack with 8 BS. ${\rm Time}_{tr}$ and ${\rm N}_T$ denote training time and frame numbers of temporal information.}
    \label{tab:speed}
    \centering
    \small
    \resizebox{0.9\linewidth}{!}{
    \setlength{\tabcolsep}{0.5mm}{
    \begin{tabular}{ccccccc}
    \toprule[1pt]
        Tracker (BS) & FLOPs&Params & FPS &Mem & $\rm Time_{tr}$ &$N_{T}$\\\midrule[1pt]
         OSTrack (128)&31.2G&92M&94&35G&38h&1\\
         +\textbf{MPT} (128)&35.9G&105M&86&6G&5h&30\\
         \hline
         ARTrack (8)&55.2G&202M&26&50G&$26+53h$&7\\
         +\textbf{MPT} (128)&56.8G&215M&26&21G&12h&30\\
            \bottomrule[1pt]
    \end{tabular}}}
    \vspace{-1.5em}
\end{table}

In addition, compared with some trackers like ODTrack and ARTrackV2, which adopt sequential-level training or multi-template strategies to enhance their tracking performance, our MPT requires less memory and computing resources during training. As shown in Tab.~\ref{tab:speed}, taking a sequential-training tracker (ARTrack-B256) as a reference, the setting of 8 batch sizes (BS) needs 50G GPU memory, within more than 3 days of training. 
In contrast, our MPT adopts lightweight architecture and efficient frame-level training, introducing only 13M parameters (Params) and negligible speed (FPS) degradation for baselines. For OSTrack with 128 BS, our MPT only needs 5 hours of fine-tuning using 6G memory. It demonstrates that our MPT is training efficient, and offers incremental benefits to various trackers with minimal training resources.

\begin{table}[t]
    \caption{\textbf{Impact of positional encodings,} including respective role of each PE and comparisons of TPE variants.}
    \label{tab:pe}
    \centering
    \resizebox{\linewidth}{!}{
    \setlength{\tabcolsep}{0.5mm}{
    \begin{tabular}{cc|ccc|ccc}
    \toprule[1pt]
    &\ding{172}&\ding{173}&\ding{174}&\ding{175}&\ding{176}&\ding{177}&\ding{178}\\
    &Ours&w/o SPE&w/o TPE&w/o P&$\rm Init_{ran}$&$\rm Init_{lin}$&Non-Learn\\\midrule[1pt]
    $\rm LaSOT_{\rm EXT}$&48.7&44.8&47.5&48.1&45.9&47.4&47.4\\
    VOT2022&0.572&0.556&0.547&0.573&0.567&0.561&0.552\\
    $\Delta$&$-$&$-2.8\%$&$-1.9\%$&$-0.3\%$&$-1.7\%$&$-1.2\%$&$-1.7\%$\\
    \bottomrule[1pt]
    \end{tabular}
    \vspace{-1.0em}
    }}
\end{table}

\subsection{Ablation Study and Analysis}
\label{sec:ablation_study}
To deeply analyze our MPT, we conduct comprehensive ablation studies on $\rm LaSOT_{EXT}$ and VOT2022, choosing OSTrack-B256+MPT as our baseline (denoted as Ours). We present AUC and EAO for two benchmarks, respectively, and also the averaged performance change (denoted as $\Delta$).

\noindent \textbf{Positional encoding.} 
To explore impacts of the adopted three positional encodings, we individually remove one type of PE, shown as Tab.~\ref{tab:pe}. In our motion encoder, the spatial PE (SPE) is mainly used to encode each point coordinate to the visual embedding space, facilitating the alignment of motion and visual features in terms of spatial positions. When we replace our SPE with a simple MLP (\ding{173}), the tracking performance decreases by 2.8\% on average.
Besides, our temporal PE (TPE) integrates temporal orders to motion tokens in a non-linear way, and the employed point embeddings (P) identify the type of corner points for motion tokens. Removing any of them (\ding{174} and \ding{175}) results in a certain degree of performance decline. These results indicate that each type of PE plays a crucial role in the performance of our MPT, particularly SPE and TPE.

Furthermore, we adopt a non-linear initialization on a learnable embedding for our TPE (formulated as Eq.~\ref{eq:temproal_pe}) to make the model dynamically focus on the significant motion information. To explore the impact of different TPE initializations, we compare random (\ding{176}) and linear (\ding{177}) initializations with our non-linear method(\ding{172}). It shows that our non-linear TPE initialization performs superior to others. Besides, to investigate the efficacy of the learnable attribute in our TPE, we also compare our method with a variant with non-learnable TPE (\ding{178}). Disabling the learning capability of TPE results in a performance degradation of 1.7\%.


\noindent \textbf{Loss function.} We formulate three loss terms for model optimization. In addition to the baseline tracking loss $\mathcal{L}_{\rm Tr}$, we introduce the other two loss terms to enhance supervision. Among them, the motion regression loss $\mathcal{L}_{\rm M}$ guides Rep. tokens to extract essential motion patterns from the trajectory. The adaptive weight loss $\mathcal{L}_{\rm W}$ is employed to explicitly supervise the weight prediction, allowing the model to adaptively adjust its reliance on visual or motion cues. Results reported in Tab.~\ref{tab:loss} demonstrate the significance of both $\mathcal{L}_{\rm M}$ and $\mathcal{L}_{\rm W}$. The combined utilization of all three losses (Ours) boosts the model to achieve superior performance.

\begin{table}[t]
    \caption{\textbf{Impact of different losses and components.}}
    \label{tab:loss}
    \centering
    \small
    \resizebox{\linewidth}{!}{
    \setlength{\tabcolsep}{0.5mm}{
    \begin{tabular}{cc|cc|ccc}
    \toprule[1pt]
    &Ours&$\mathcal{L}_{\rm Tr}+\mathcal{L}_{\rm M}$&$\mathcal{L}_{\rm Tr}$&Motion&Vision&w/o Weight\\ \midrule[1pt]
    $\rm LaSOT_{\rm EXT}$&48.7&48.6&45.1&48.5&46.9&46.4\\
    VOT2022&0.572&0.558&0.556&0.525&0.530&0.568\\
    $\Delta$&$-$&$-0.8\%$&$-2.6\%$&$-2.5\%$&$-3.0\%$&$-1.4\%$\\
    \bottomrule[1pt]
    \end{tabular}
    \vspace{-1.5em}
    }}
\end{table}

\begin{figure}[t!]
    \centering
    \includegraphics[width=0.95\linewidth]{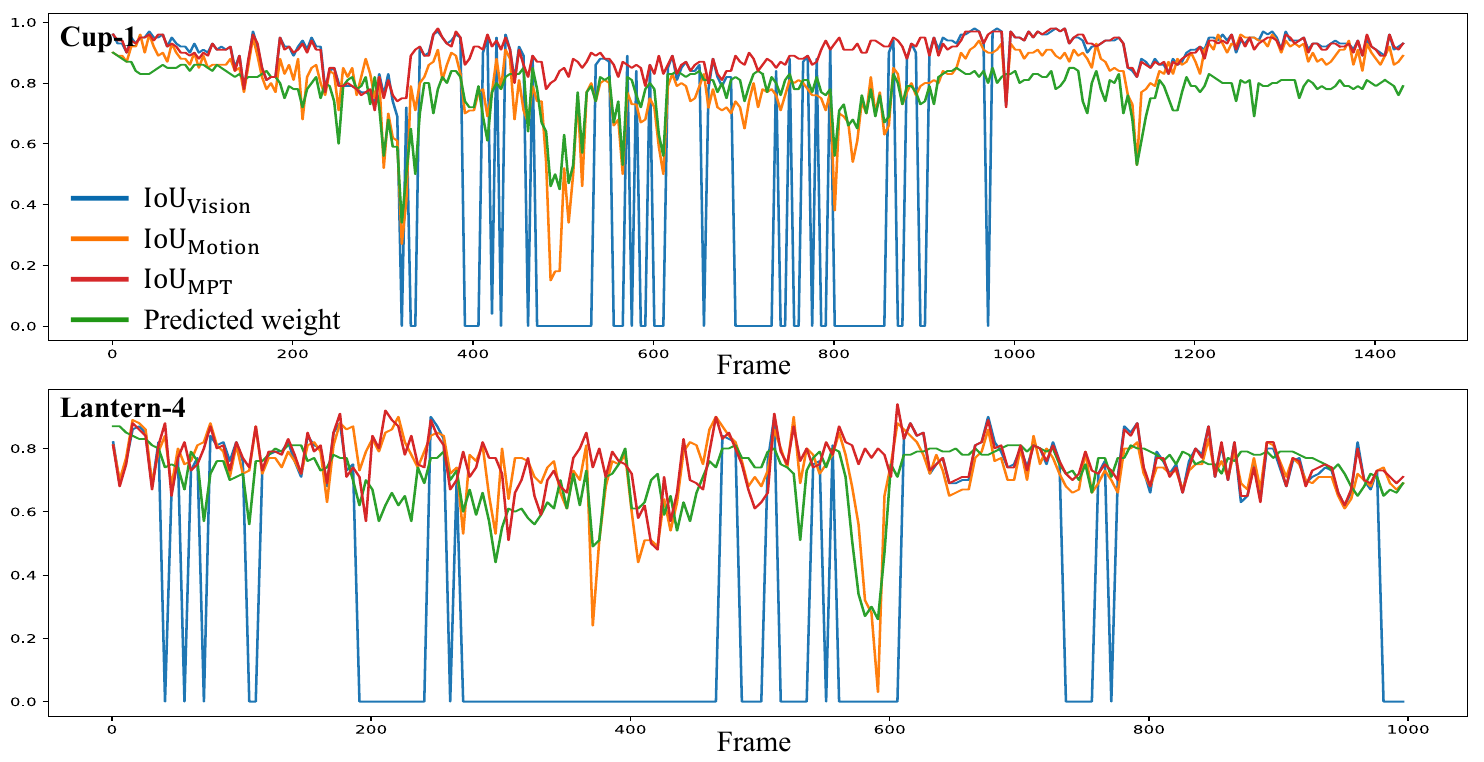}
    \vspace{-0.5em}
    \caption{\textbf{Illustration of adaptive weights, and tracking performance based on different cues.}}
    \vspace{-0.5em}
    \label{fig:weight}
\end{figure}

\noindent \textbf{Adaptive weight mechanism.} Results in Tab.~\ref{tab:loss} (right) demonstrate the respective roles of visual and motion inputs, as well as the effectiveness of the proposed adaptive weight mechanism. ``Motion" represents the tracking results directly predicted by the motion head $\rm Head_M$, while ``Vision" shows the results of the baseline vision-based tracker. We also implement a variant of MPT without the adaptive weight mechanism, denoted as ``w/o Weight". That is, instead of calculating the weighted average with the original visual features, we directly use the output of the fusion decoder for tracking predictions. As observed, our MPT, compared to the tracker relying solely on visual information, yields more robust tracking results by incorporating motion prompts.  
Additionally, the adaptive weight mechanism contributes to our method achieving further improvements.

Besides, we visualize frame-level qualities of tracking results predicted by the aforementioned three variant models, namely ``Vision", ``Motion", and our MPT. As shown in Fig.~\ref{fig:weight}, when encountering plenty of distractors, the prediction of the vision model (blue curve) is notably unstable. In contrast, our MPT (red curve) can exhibit superior and robust performance. Furthermore, the values of our dynamic weights (green curve) are consistent with the quality of motion predictions (orange curve). This demonstrates that our adaptive weight mechanism can achieve a dynamic balance between visual and motion cues.

\begin{figure}[t!]
    \centering
    \includegraphics[width=0.95\linewidth]{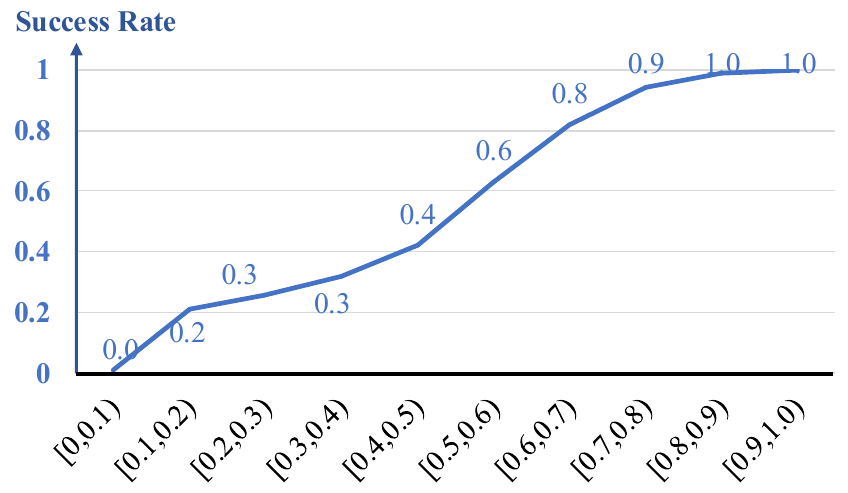}
    \vspace{-0.5em}
    \caption{\textbf{Success rate of among varying trajectory qualities.}}
    \vspace{-0.5em}
    \label{fig:sensitivity}
\end{figure}

\begin{figure}[t!]
    \centering
    \includegraphics[width=0.95\linewidth]{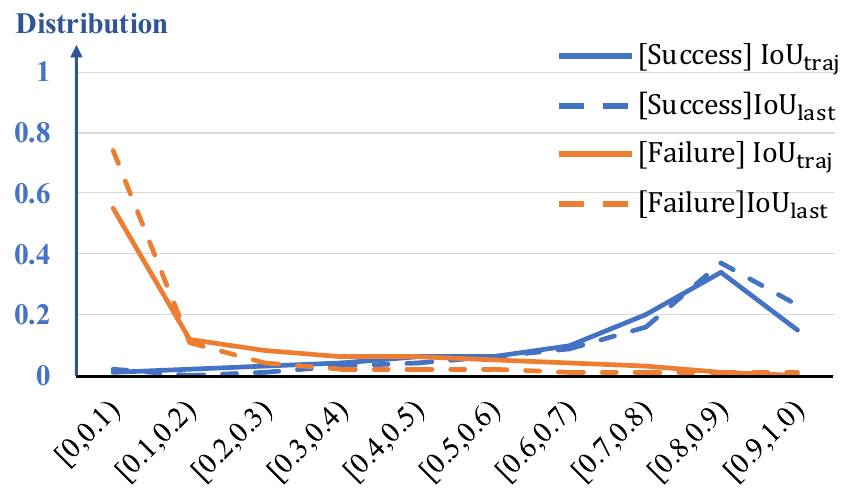}
    \vspace{-0.5em}
    \caption{\textbf{IoU distributions of average trajectory and the last frame in success and failure cases.}}
    \vspace{-0.5em}
    \label{fig:case}
\end{figure}

\noindent \textbf{Sensitivity to the trajectory quality.} To further analyze the sensitivity of our MPT to varying noisy trajectories, we evaluate the success rate ($\rm IoU > 0.5$) under different quality trajectory inputs on $\rm LaSOT_{\rm EXT}$, as shown in Fig.~\ref{fig:sensitivity}. Here, [a,b) represents the range where the average IoU of the trajectory is no less than a and less than b. We can find that the success rate increases significantly with the improvement of trajectory quality, especially when the trajectory IoU is in the range of [0.6, 1.0). Besides, our method shows a certain level of robustness to noisy trajectory inputs, obtaining a success rate of 21\% even when the trajectory IoU is in the range of [0.1, 0.2).

In addition, to deeply explore the success and failure cases of our MPT, we separately evaluate the distribution of trajectory IoU ($\rm IoU_{traj}$) and tracking IoU of the last frame ($\rm IoU_{last}$), in both conditions on $\rm LaSOT_{\rm EXT}$.
Here, we define a success case when the predicted IoU of our MPT exceeds that of the visual baseline by more than 0.3. Conversely, when the predicted IoU of our MPT is lower than that of the visual baseline by more than 0.3, we define it as a failure case. 
As shown in Fig.~\ref{fig:case}, in success cases, the input trajectory IoU primarily falls within a relatively broader range of [0.6, 1.0), especially in [0.8, 0.9). In failure cases, the input trajectory IoU mainly falls within the range of [0.0, 0.2), particularly in [0.0, 0.1). This indicates that addressing misleading tracking issues caused by extremely poor-quality trajectories remains a challenge for our method. This is a limitation of our method that requires future research.


\begin{table}[t]
    \caption{\textbf{Impact of fine-tuning strategies,} in terms of tracking performance and training efficiency.}
    \label{tab:finetune}
    \centering
    \resizebox{0.85\linewidth}{!}{
    \setlength{\tabcolsep}{0.5mm}{
    \begin{tabular}{ccccc}
    \toprule[1pt]
    &OSTrack&Ours&FinetuneAll&FinetuneHead\\
    \midrule[1pt]
    $\rm LaSOT_{\rm EXT}$&46.9&48.7&48.3&48.1\\
    VOT2022&0.530&0.572&0.567&0.548\\
    Mem (G)&35.0&6.1&37.5&7.7\\
    $\rm Time_{tr}$ (h)&38&5&8&6\\
    \bottomrule[1pt]
    \end{tabular}
    \vspace{-2.0em}
    }}
\end{table}

\noindent \textbf{Fine-tuning strategies.} Our method freezes all of the baseline parameters, and only fine-tunes parameters of MPT during training. We also evaluate two variants, shown in Tab.~\ref{tab:finetune}. FinetuneAll fine-tunes both MPT and baseline parameters, while FinetuneHead fine-tunes parameters of MPT and the tracking head of the baseline, \ie, only freezing the backbone of the baseline model. Our method slightly outperforms the other two variants, and is more efficient in terms of memory footprint and training time. The potential reason is that freezing the baseline parameters enables the model to focus on learning motion cues and the fusion mechanism.

\begin{figure}[tbp]
    \centering
    \includegraphics[width=\linewidth]{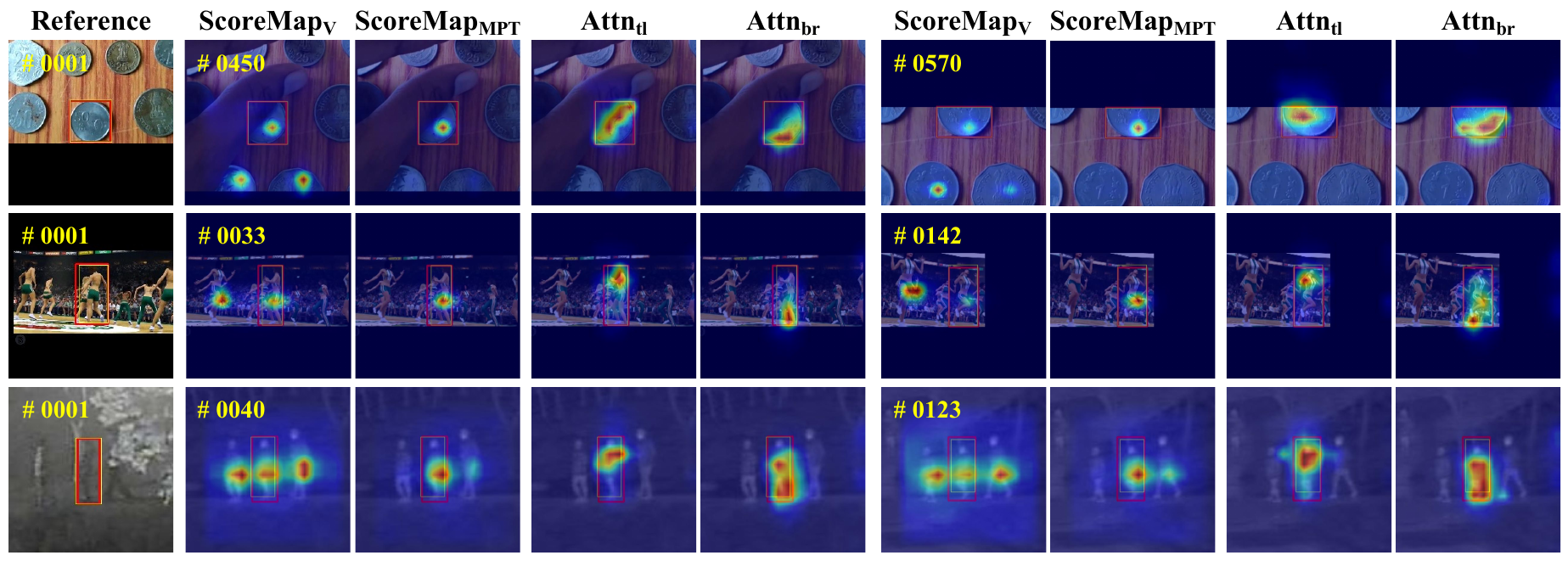}
    \caption{\textbf{Visualization of cross-attention maps, and discriminative comparison of score maps.}}
    \vspace{-0.5em}
    \label{fig:attn}
\end{figure}

\noindent \textbf{Visualization of cross-attention maps.} For a more straightforward understanding of how our MPT fuses visual and motion features, we visualize the cross-attention maps (averaged over heads) of the last decoder block, shown as Fig.~\ref{fig:attn}. Among them, ``$\rm Attn_{tl}$" and ``$\rm Attn_{br}$" represent cross-attention maps between visual features and two representative motion tokens ($\rm RT_{tl}$ and $\rm RT_{tl}$), respectively. In addition, we also visualize the score maps predicted by the vision-based baseline tracker ($\rm ScoreMap_V$), and our MPT model ($\rm ScoreMap_{MPT}$). We find that it is difficult to distinguish distractors based on a solely visual model. Due to the complement of motion prompts, our MPT pays more attention to the top-left and bottom-right corner points of the interested target. This enables the model to weaken the response to distractors, thereby achieving robust tracking.

Due to space limitations, more visualizations can be found in Appendix~\ref{sec:vis}, including the optimizing process during training (Fig.~\ref{fig:loss}), qualitative comparisons (Fig.~\ref{fig:bbox}), and some failure cases (Fig.~\ref{fig:bbox_fail}).

\noindent\textbf{Limitations.} Although our MPT excels in training efficiency and tracking robustness, its tracking accuracy is slightly inferior to those sequential training methods. We will further explore effective methods to capture dynamic relations in efficient prompting. In addition, as shown in Fig.~\ref{fig:case}, addressing misleading tracking issues caused by extremely poor-quality trajectories remains a challenge for our method. which requires future research.

\section{Conclusion}
In this paper, we present a flexible and efficient motion prompt tracking method. It can complement existing vision-based trackers by learning motion prompts, thereby maintaining robust tracking in challenging scenarios. 
Our method includes a motion encoder with three different positional encodings, a Transformer-based fusion decoder, and an adaptive weight mechanism. 
Motion trajectories of the object are independently encoded and then adaptively fused with visual features. 
We integrate our method into five visual tracking models. Experiments on seven datasets show that our method efficiently enhances tracking performance within minimal computational costs and memory usage. 
We hope that our new insights into temporal tracking and prompt tracking will leverage future research on these topics.


\section*{Acknowledgments} The computational resources were provided by the National Academic Infrastructure for Supercomputing in Sweden (NAISS) at C3SE partially funded by the Swedish Research Council grant 2022-06725. This work was supported in part by the Wallenberg Artificial Intelligence, Autonomous Systems and Software Program (WASP), funded by Knut and Alice Wallenberg Foundation, in part by the strategic research environment ELLIIT funded by the Swedish government, and the Swedish Research Council grant 2022-04266, in part by National Natural Science Foundations of China (No. 62402084 and No. 62176041), in part by Liaoning Province Science and Technology Plan (No. 2024JH2/102600040), and in part by China Postdoctoral Science Foundation (No. 2024M750319).

\section*{Impact Statement} This work advances the field of visual object tracking through efficient motion prompt learning. The technique may benefit real-time applications such as surveillance systems.
\bibliography{main}
\bibliographystyle{icml2025}

\newpage
\appendix
\onecolumn
\section*{Appendix}
This appendix contains proof of non-linear temporal positional encoding in Section~\ref{sec:nontpe}, additional implementation details in Section~\ref{sec:detail}, extended ablation studies in Section~\ref{sec:ext_ablation}, in-depth performance analysis in Section~\ref{sec:performance}, and more visualizations in Section~\ref{sec:vis}.

\section{Proof of Non-Linear Temporal Positional Encoding}
\label{sec:nontpe}
\begin{figure}[ht]
    \centering
    \includegraphics[width=0.8\linewidth]{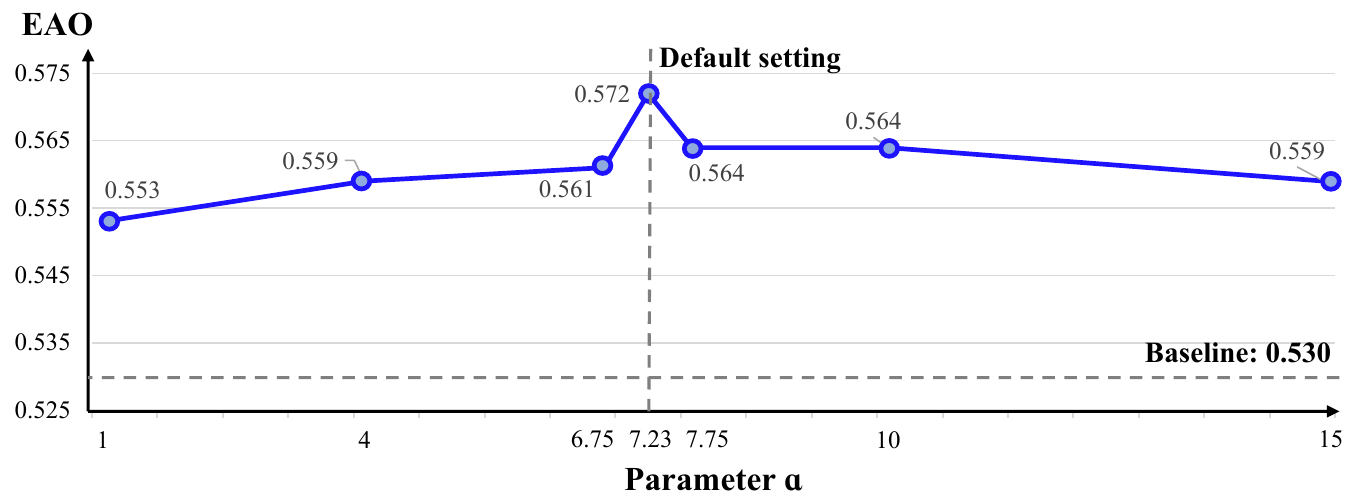}
    \caption{Impacts of different parameter $\alpha$ on VOT2022.}
    \label{fig:alpha}
\end{figure}
To allow our model to perceive temporal orders of motion tokens, we adopt a learnable temporal positional encoding and initialize it in the following non-linear manner:
\begin{equation}
\label{eq:temproal_pe2}
    \begin{split}
    {\rm PE}(t,2i) &=\sin\left(\frac{\alpha \lg(t+1)}{n^{2i/d}}\right), \\
    {\rm PE}(t,2i+1) &=\cos\left(\frac{\alpha \lg(t+1)}{n^{2i/d}}\right).
    \end{split}
\end{equation}
The parameter $\alpha$ is employed to control the frequency range, and set to 7.23 based on the Nyquist frequency~\cite{nyquist}. The proof of this setting is presented as follows.

According to the theory of Nyquist frequency, a signal should be sampled at a rate at least twice its highest frequency to avoid aliasing.
Due to the logarithmic mapping of the time parameter $t$, the frequency $\omega$ of Eq.~\ref{eq:temproal_pe2} changes in a non-linear way and needs to be determined using the derivative (note that $\lg(t+1)=\ln(t+1)/\ln(10)$):
\begin{equation}
\omega(t,i)=\frac{\partial_t PE(t,2i)}{PE(t,2i+1)}=\frac{\alpha}{\ln(10) (t+1)n^{2i/d}}.
\end{equation}
Its maximum is attained at $(t,i)=(0,0)$. From $\omega(0,0)=\pi$, we obtain $\alpha=\pi\ln(10)\approx7.23$. 

Moreover, we also verify its experimental effectiveness by comparing it with other parameter values. As shown in Fig.~\ref{fig:alpha}, results on VOT2022~\cite{vot2022} show that the best performance is attained at $\alpha=7.23$. Other parameter values cause slight performance decays, but they are still superior to the baseline tracker (marked as the horizontal dashed line).

\section{Implementation Details}
\label{sec:detail}
\noindent \textbf{Model.} For our motion encoder, we encode each 2-dimensional coordinate into a motion token with the same dimension as the baseline visual features, which are 768 and 256 for OSTrack-B / ARTrack-B, and SeqTrack-B, respectively. The lightweight fusion decoder is implemented as a two-layer Transformer network. For each layer, the number of attention heads is 8, and the hidden size of MLP is set to 1024 and 256 for OSTrack / ARTrack and SeqTrack, respectively. The weight head $\rm Head_W$ and motion head $\rm Head_M$ are both implemented by a two-layer MLP, where the hidden size is 256. 

\begin{figure}[ht]
    \centering
    \includegraphics[width=0.8\linewidth]{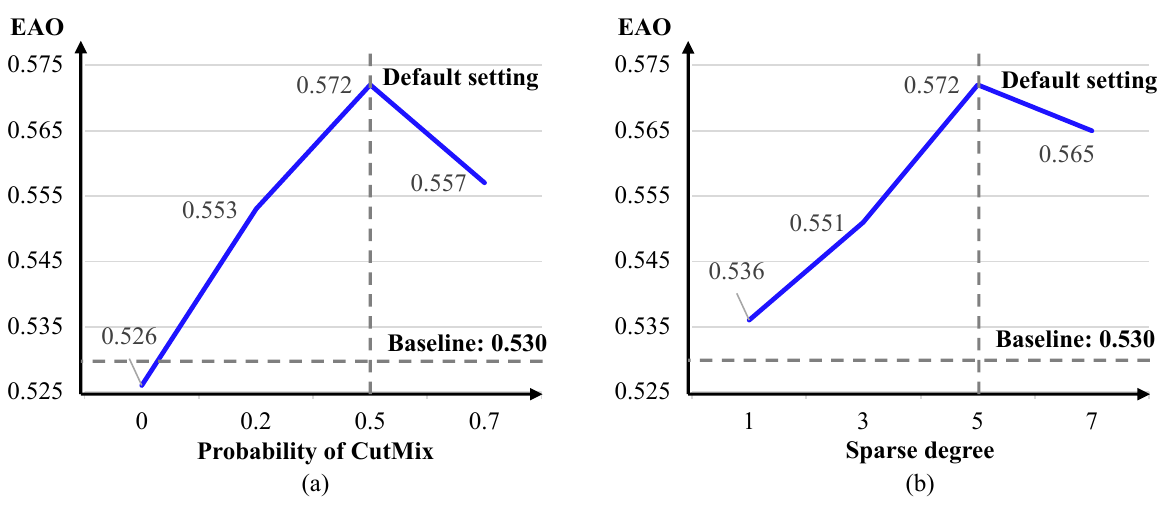}
    \caption{Impacts of data augmentations on VOT2022.}
    \label{fig:aug}
\end{figure}

\noindent \textbf{Training.} The model is trained for 60 epochs with 60k image pairs per epoch. We set the batch size to 128, and the learning rate is decreased by a factor of 10 after 40 epochs. The initial learning rate and other training settings are set the same as corresponding baseline trackers.
For the motion input, we sparsely sample a $T$-frames trajectory either before or after the current search frame. In a trajectory, we employ equally spaced sparse sampling to preserve the original motion pattern of the target. 
Such an offline trajectory generation strategy not only simplifies and accelerates the training process, but also allows the model to confront real tracking noises and learn to adapt to motion trajectories with varying qualities. In addition, we adopt CutMix strategy~\cite{yun2019cutmix} on search images to simulate challenges like background clutter. Specifically, we crop the object area from another frame, and paste it on a random position of the search image. This data augmentation offers more chances for our model to learn how to effectively fuse and utilize extra motion information. The following Section will analyze the impacts of the adopted data augmentations.

\begin{figure}[htbp]
    \centering
    \includegraphics[width=0.5\linewidth]{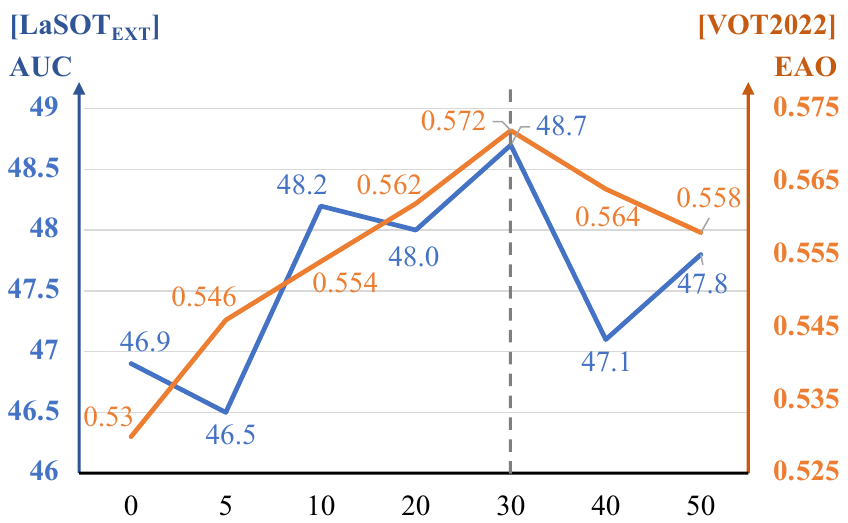}
    \caption{Impact of the trajectory length. Our MPT obtains the best performance when the length of trajectories $T=30$.}
    \label{fig:length}
\end{figure}

\section{Extended Ablation Study}
\label{sec:ext_ablation}
\subsection{Analysis of Data Augmentations}
Fig.~\ref{fig:aug} illustrates the individual impacts of CutMix and sparse sampling.
For the CutMix, training our method without using CutMix strategy (\ie probability is set to 0) obtains slightly inferior performance than the baseline visual tracker. The potential reason is that most original training samples are easy for the baseline visual model, limiting chances for our motion model to learn the ability to complement and optimize tracking predictions. CutMix, however, can simulate hard scenes by introducing distractors and slight occlusion, which mitigates this training issue. Fig.~\ref{fig:aug}(a) shows that the best performance of our model is attained when the probability of CutMix is set to 0.5. Therefore, we set the probability of 0.5 as the default in our experiments.

For sparse sampling, increasing the range of sparseness for trajectory sampling enables the model to learn from more diverse motion inputs, but it concurrently means that our model faces more intricate and challenging motion patterns during the training. As shown in Fig.~\ref{fig:aug}(b), the sparseness of 5 is an optimized choice, which is also our default setting.

\subsection{Impact of Trajectory Length} 
Fig.~\ref{fig:length} illustrates the impact of the motion trajectory length $T$ on the performance of our MPT. Our method exhibits improved performance as $T$ increases, and reaches its peak at $T=30$. Since the longer trajectory might bring more motion noises, further increasing the trajectory length results in slight performance degradation. Therefore, we set $T=30$ in our experiments. This ablation study confirms our conjecture that long-term temporal information can provide richer tracking cues. Different from prior works~\cite{kim2022sequential,wei2023autoregressive} employing heavy temporal processing mechanisms, our lightweight MPT independently encodes motion inputs using negligible computing resources, which allows the model to manage long-term motion trajectories.

\begin{table}[tbp]
\centering
\caption{\textbf{Quantitative analysis of performance bias}}
\label{tab:bias}
\begin{tabular}{cc cc|cc|cc|cc}
\toprule
   &&\multicolumn{2}{c|}{LaSOT}& \multicolumn{2}{c|}{$\rm LaSOT_{\rm EXT}$}&   \multicolumn{2}{c|}{TNL2K}&  \multicolumn{2}{c}{ALL}\\ 
     &     & Num &Gain  & Num &Gain& Num &Gain& Num &Gain\\ \midrule
\multirow{2}{*}{OSTrack-B256}&\ding{172}  &50&+4.8&80&+4.8&284&+2.3&414&+4.0\\
&\ding{173}&230&-0.7&70&-1.6&416&-1.4&716&-1.2\\ \midrule 
\multirow{2}{*}{OSTrack-B384}&\ding{172}  &40&+5.6&71&+4.5&252&+2.1&363&+4.1\\
&\ding{173}&240&-1.3&79&-0.5&448&-0.7&767&-0.8\\ \midrule
\multirow{2}{*}{SeqTrack-B256}&\ding{172}&44&+5.5&70&+2.8&271&+3.8&385&+4.0\\
&\ding{173}&236&-0.2&80&-0.4&429&-0.4&745&-0.3\\ \midrule
\multirow{2}{*}{SeqTrack-L384}&\ding{172}&38&+9.3&76&+4.4&234&+2.8&348&+5.5\\
&\ding{173}&242&+0.2&74&-1.8&466&-0.3&782&-0.6\\ \midrule
\multirow{2}{*}{ARTrack-B256}&\ding{172}&43&+5.9&78&+4.8&255&+3.2&376&+4.6\\
&\ding{173}&237&-1.9&72&-0.9&445&-0.8&754&-1.2\\ 
\bottomrule
\end{tabular}
\end{table}

\section{In-depth Performance Analysis}
\label{sec:performance}

\subsection{In-depth Analysis about Performance Bias}

Aiming to reduce tracking failures by motion cues, our MPT can significantly improve tracking robustness over accuracy. Our advantages are more evident in hard scenarios (like VOTs, $\rm LaSOT_{EXT}$). This may cause a slight bias in the overall performance across different trackers or benchmarks. To give evidence for this conjecture, we dynamically split each dataset into two subsets based on performance of baselines, namely hard / easy sets. As shown in Tab.~\ref{tab:bias}, the hard set (\ding{172}) includes hard videos with baseline AUCs below 50.0\%, while the easy set (\ding{173}) includes the remaining videos. Hard set proportions of LaSOT, TNL2K, $\rm LaSOT_{EXT}$ are around 15\%, 30\%, 50\%. This aligns with the trend in our average performance gains: 0.4\%, 0.6\%, 1.6\%. 
Moreover, our MPT shows better consistency in hard / easy sets, achieving average AUC changes of +4.4\% / -0.8\%, with small standard deviations of 0.6 / 0.4 across five baselines.

\begin{figure}[htbp]
    \centering
    \includegraphics[width=0.6\linewidth]{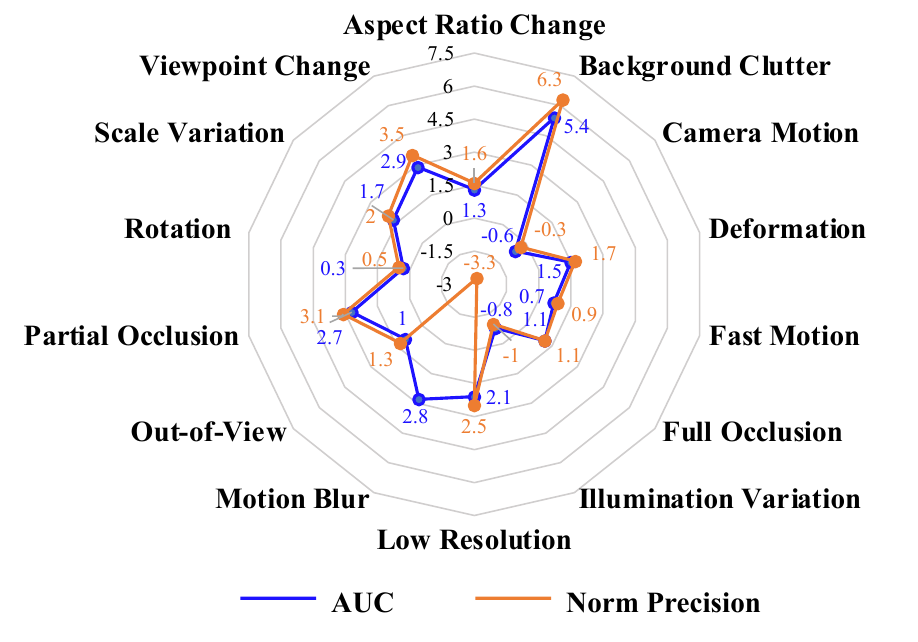}
    \caption{Performance gains of our method on each attribute.}
    \label{fig:attr}
\end{figure}
\subsection{Attribute Analysis}

For a more intuitive observation of the impacts of our MPT on different attributes, we report performance gains achieved by our MPT across different attributes on $\rm LaSOT_{EXT}$~\cite{fan2021lasotext}, shown as Fig.~\ref{fig:attr}. As we can see, our MPT significantly enhances the baseline tracker when faced with the challenge of background clutter, obtaining 6.3\% AUC improvements. Besides, challenges like viewpoint change, partial occlusion, low resolution, deformation, and full occlusion can also be better addressed by our MPT.

\section{Visualizations}
\label{sec:vis}
\subsection{Visualization of the Training Process}
\label{sec:training}
\begin{figure*}[htbp]
    \centering
    \includegraphics[width=\linewidth]{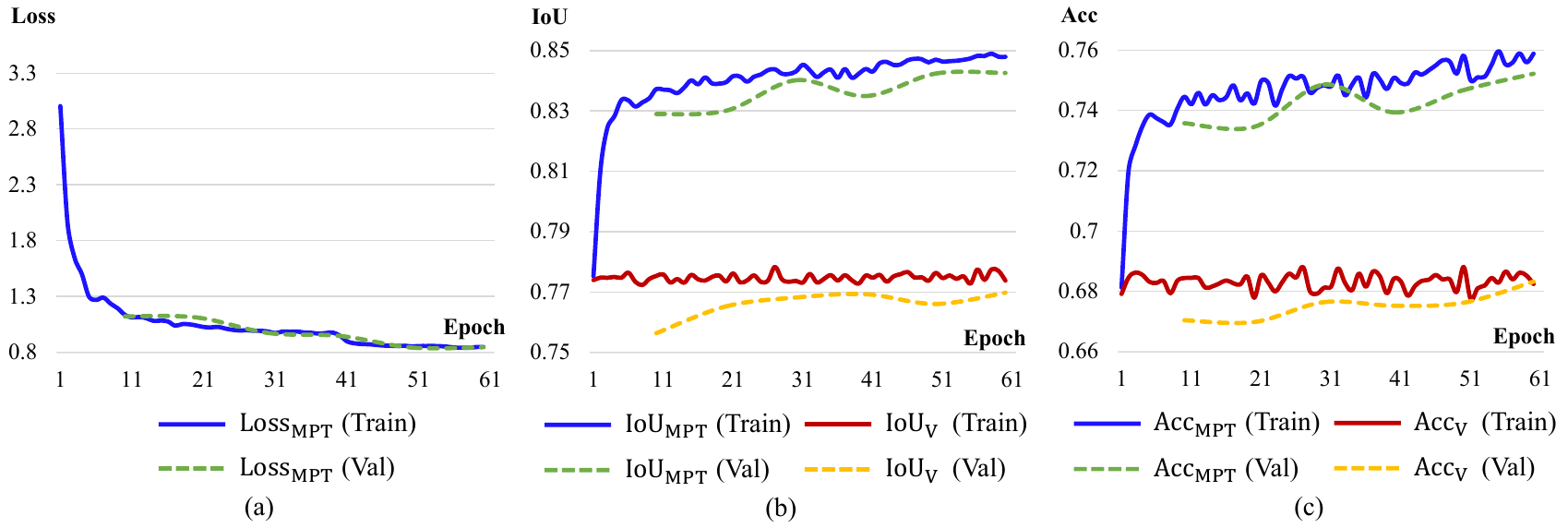}
    \caption{Visualization of optimizing process of our model during training.}
    \label{fig:loss}
\end{figure*}

As shown in Fig.~\ref{fig:loss}, we present the optimizing process of our model by visualizing dynamic changes of (a) loss values, (b) Intersection-over-Union (IoU), and (c) location accuracy of the score map (Acc). 
Here, we consider the location prediction to be accurate if the position of the maximum value in the score map coincides with the center of the target groundtruth.
As we can see, due to fine-tuning few parameters, the model exhibits fast convergence during the training process. Moreover, compared with the baseline visual model OSTrack-B256~\cite{ye2022ostrack}, our motion prompt-based model yields a more accurate score map (see Fig.~\ref{fig:loss}(c)), consequently leading to a more precise prediction of the object position (see Fig.~\ref{fig:loss}(b)). These curves further substantiate the efficacy of our MPT method in terms of both training efficiency and performance.

\begin{figure*}[tb]
    \centering
    \includegraphics[width=0.8\linewidth]{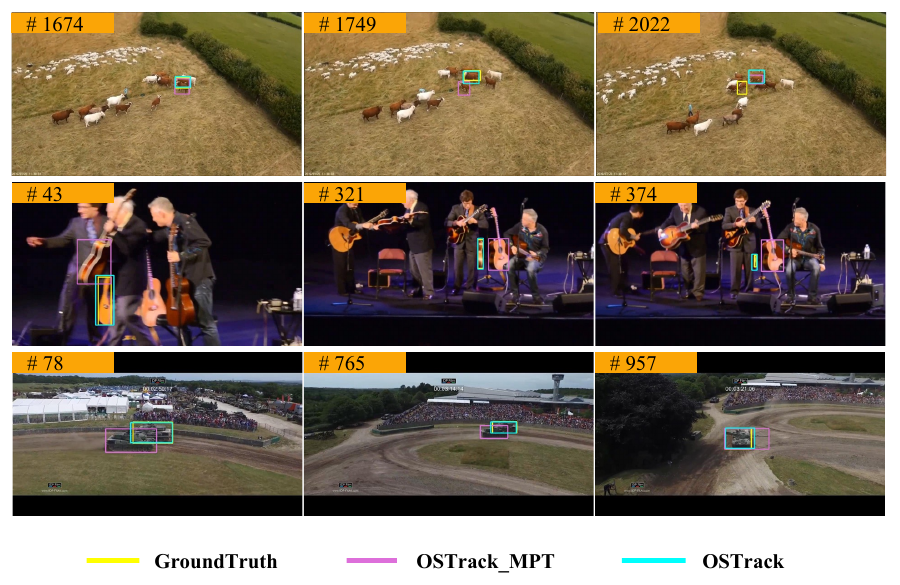}
    \caption{Several failure cases of our method (``\_MPT").}
    \label{fig:bbox_fail}
\end{figure*}

\begin{figure*}[ht]
    \centering
    \includegraphics[width=0.75\linewidth]{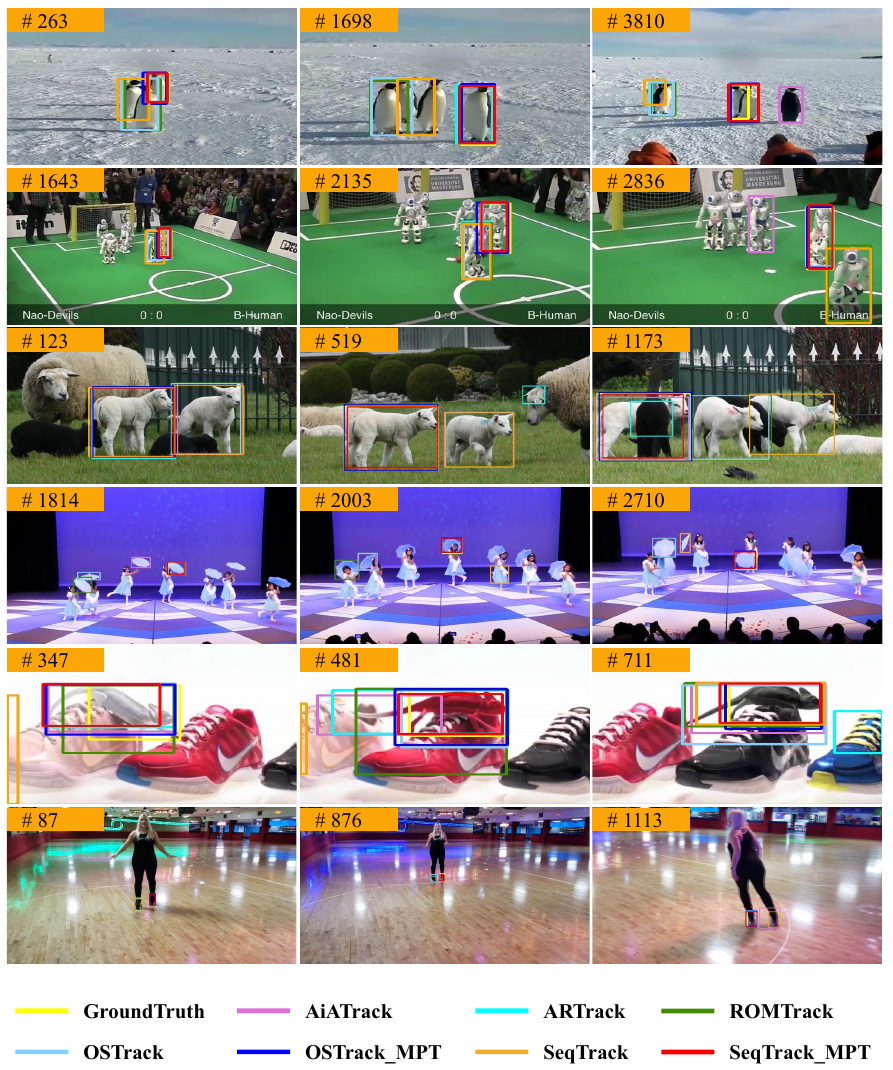}
    \caption{Qualitative comparisons of our trackers (``\_MPT") with other SOTA trackers.}
    \label{fig:bbox}
\end{figure*}

\subsection{Qualitative Comparison}
\label{sec:qualitative}
To intuitively demonstrate the effectiveness of our MPT, we make a qualitative comparison of our methods with corresponding baseline trackers and other state-of-the-art (SOTA) trackers. As shown in Fig.~\ref{fig:bbox}, our MPT allows SOTA vision-based baseline trackers to perform more robustly when faced with challenges, like distractors, occlusion, severe appearance changes, fast motion, etc.

Moreover, we make a video (please find it at \url{https://github.com/zj5559/Motion-Prompt-Tracking}) to show more qualitative comparisons on three challenging sequences. For each sequence, we visualize bounding boxes predicted by our method (``+MPT'') and the corresponding baseline tracker OSTrack. We also compare score maps output by these two trackers. The historical trajectories are plotted in our score maps (``Feat\_MPT''). Compared with the baseline tracker, our MPT can utilize object motion cues to help the tracker recognize the interested object from multiple distractors. Besides, when faced with severe occlusion, our MPT can predict the next position of the object based mainly on the motion prompt, thereby avoiding tracking drift.

However, since motion trajectories often include inherent noises like wrong observations, low-quality motion cues sometimes interfere with visual cues. We propose an adaptive weight mechanism to mitigate this phenomenon, but some tracking failures still inevitably happen. Some failure cases of our method are shown in Fig.~\ref{fig:bbox_fail}.

\end{document}